\documentclass[11pt]{article}

\usepackage[final]{acl}

\usepackage{times}
\usepackage{latexsym}
\usepackage[T1]{fontenc}
\usepackage[utf8]{inputenc}

\usepackage{microtype}
\usepackage{inconsolata}
\usepackage{booktabs}
\usepackage{amsmath}
\usepackage{amssymb}
\usepackage{amsfonts}
\usepackage{graphicx}
\usepackage{subcaption}
\usepackage{algorithm}
\usepackage{algpseudocode}
\usepackage{xcolor}
\usepackage{tcolorbox}
\usepackage{multirow}
\usepackage{array}
\usepackage{makecell}
\usepackage{colortbl}

\definecolor{phcolor}{rgb}{0.8, 0.2, 0.0}
\definecolor{notecolor}{rgb}{0.0, 0.4, 0.0}

\newcommand{\Hnumber}[1]{#1}

\newcommand{\impNoVetoErrSB}{21}
\newcommand{\impNoVetoErrDA}{14}
\newcommand{\impGatedErrSB}{15}
\newcommand{\impGatedErrDA}{6}
\newcommand{\impDeltaErrSB}{6}
\newcommand{\impDeltaErrDA}{8}

\newcommand{\impTokSaveSB}{14}
\newcommand{\impTokSaveDA}{19}
\newcommand{\impNoVetoSuccSB}{49}
\newcommand{\impNoVetoSuccDA}{64}
\newcommand{\impGatedSuccSB}{44}
\newcommand{\impGatedSuccDA}{63}
\newcommand{\impDeltaSuccSB}{5}
\newcommand{\impDeltaSuccDA}{1}
\newcommand{\impRecallSB}{0.72}
\newcommand{\impRecallDA}{0.69}
\newcommand{\impFvrSB}{0.14}
\newcommand{\impFvrDA}{0.16}

\DeclareMathOperator*{\argmax}{arg\,max}

\title{How to Speculate about Uncertainty in Agentic Coding?\\A Draft-Model Gate Method}

\author{
  Konstantin Grotov\thanks{\ Corresponding author.} \\
  ITMO University \\
  \texttt{konstantin.grotov@gmail.com}
  \And
  Valentin Malykh \\
  MWS AI, IITU University, \\
  Trusted AI Research Center, RAS \\
  \texttt{valentin.malykh@phystech.edu}
}

\author{
  Konstantin Grotov\thanks{\ Corresponding author.} \\
  Tel Aviv University \\
  Tel Aviv, Israel \\
  \texttt{konstantin.grotov@gmail.com}
  \And
  Valentin Malykh \\
  IITU \\
  Almaty, Kazakhstan \\
  \texttt{valentin.malykh@phystech.edu}
}

\begin{document}

\maketitle

\begin{abstract}
LLM agents deployed for software engineering fail expensively: they act
\emph{confidently wrong}, and bad actions are recognized only after costly
execution and retry. We present \textbf{Speculative Uncertainty (SU)}, a
method that recovers a predictive failure signal for a black-box
agent from its output tokens alone, with no access to logits, weights,
activations, or repeated sampling. Inverting speculative decoding, a small open-weight
\emph{draft} model scores the agent's already-generated trajectory in a single
forward pass. From these speculative cross-likelihoods we extract phase-aware
features by separating the reasoning and action spans, and calibrate them against
a verifiable objective. SU produces a failure-likelihood score that any downstream policy, such as routing, human intervention, or extra test-time compute, can consume directly.
To show the signal is actionable, we instantiate one such policy, a
pre-execution veto gate, on software engineering agents Qwen3-Coder-480B and
closed-source Claude~3.5~Sonnet, cutting execution error rate by
\Hnumber{6--8} percentage points and token cost by \Hnumber{14--19}\% in deployment, transferring
to out-of-distribution benchmarks without retraining, and generalizing across
agent models.
\end{abstract}

\section{Introduction}
\label{sec:intro}

LLM agents are increasingly deployed in production software engineering
workflows, where they autonomously draft patches, run tests, and query databases~\citep{jimenez2024swebench, badertdinov2026swe, swesmith}.
Yet behind strong benchmark scores lies an expensive failure mode: agents are
often \emph{confidently wrong}~\citep{agentic_overconfidence}. They propose an action with no signal that it could
fail, and the failure appears only when an environment rejects it---after
the code has run, the context has been filled up, and the retry has been paid for. At production scale, these silent failures inflate API cost and latency, since a rejected action is appended to the context and the agent pays for the retry — frequently looping on the same problem until the budget is exhausted~\citep{agentic_overconfidence}.

Two properties of realistic deployments constrain the space of admissible
methods. First, the most capable deployed agents are closed
API-served models exposing no logits, weights, or activations, so any method
that needs internal signals is something that we cannot use for models. Second,
sampling-based uncertainty estimators require many generations per step, which
is prohibitive at agentic horizons. Taken together, these requirements rule out most of the
uncertainty-quantification toolbox: single-turn
calibration ignores features specific to the trajectories~\citep{guo2017calibration, kuhn2023semantic},
white-box probing needs internal access, whereas sampling is too costly. The
closest prior work, Holistic Trajectory Calibration
(HTC)~\citep{zhang2025htc}, fits a calibrator on an agent's full
log-probability sequence. However, it still assumes white-box access and treats the
trajectory as a homogeneous token stream. Both assumptions break in the real
deployments we care about.

We address the access constraint with an idea inverted from speculative
decoding~\citep{leviathan2023fast, chen2023accelerating}: rather than
having a small draft model \emph{propose} tokens to accelerate a large one, we
have it \emph{score} the large agent's already-generated tokens via
teacher-forced cross-likelihood. The agent doesn't share its internal state, the
draft reads only its output text and recovers a rich uncertainty signal in a
single forward pass. We empirically observed that agent
generation has a phase structure: a high-entropy reasoning phase is
followed by a sharply lower-entropy action phase pinned down by syntax
and tool-call structure. Averaging over
the whole trajectory, as HTC does, dissolves this failure-relevant signal into
the noise of the reasoning phase, while reading the two phases apart surfaces it.

Combining these considerations, we shape the method of \textbf{Speculative Uncertainty (SU)}: a small draft
scores the agent's trajectory, phase-aware features are extracted separately
from reasoning and action spans, and a linear calibrator maps them to a probability that the next action will follow the verifiable objective. We study code execution in software engineering agents, but the approach applies to any agent
producing reasoning-then-action trajectories under a verifiable objective.

This study shows that the uncertainty of a black-box agent can be
\emph{recovered from its trajectory alone}, before the environment is ever
called, and that the recovered signal is reliable enough to act positively
in deployment. We insert a
lightweight gate between code generation and execution that predicts whether
the next action will be executed without exceptions. When failure is likely, this triggers a
cheap replan instead of an expensive execute-fail-retry cycle. On two
agentic coding benchmarks~\citep{jimenez2024swebench, dacode} the gate cuts per-call execution error rate by
\Hnumber{6-8}~percentage points and average token cost by \Hnumber{14--19}\%, with gains
transferring to an out-of-distribution benchmarks.

\paragraph{Contributions.}
\begin{enumerate}

  \item \textbf{A black-box uncertainty estimation method.} By inverting
  speculative decoding, a small draft scores the agent's generated tokens and
  recovers a usable failure signal without white-box access. This is
  exactly the regime where sampling is too costly and internals are hidden,
  making it usable with the closed-source APIs that dominate production.

  \item \textbf{Deployment guidance.} As
  one downstream policy, a pre-execution veto gate cuts per-call execution error rate by
  \Hnumber{6--8}~points and token cost by \Hnumber{14--19}\% on SWE-Bench Verified
  and DA-Code, with a tunable false-veto rate and zero-shot out-of-distribution
  transfer. We further find that a distilled \Hnumber{4B} draft is capable enough,
  the training objective barely matters, and a draft trained on one agent
  transfers to others well above the untrained baseline.
\end{enumerate}

\begin{figure*}[t]
  \centering
  \includegraphics[width=0.9\textwidth]{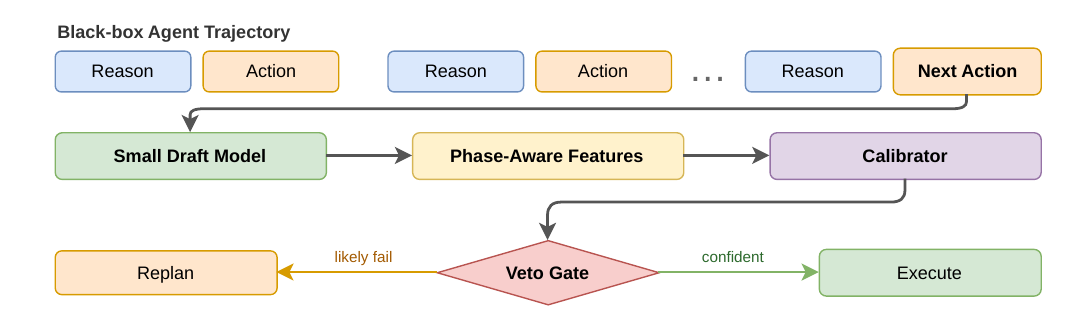}
  \caption{\textbf{The Speculative Uncertainty pipeline.} A black-box agent
    generates a trajectory of alternating reasoning and action spans, exposing
    only its output tokens. A small open-weight draft model scores these tokens
    in a single forward pass, producing per-token uncertainty signals. Features
    are extracted \emph{separately} from the reasoning and action spans, and a
    lightweight calibrator maps them to a calibrated success probability. Before
    execution, a threshold gate forwards confident actions or triggers a cheaper
    replan.}
  \label{fig:overview}
\end{figure*}

\section{Background and Related Work}
\label{sec:background}

\paragraph{Uncertainty quantification for LLMs and agents.}
Single-turn LLM calibration is well-studied~\citep{guo2017calibration, geng2024survey, kuhn2023semantic, lin2023generating}, spanning post-hoc
scaling, semantic-consistency sampling, and verbalized self-reflection. The agentic setting adds a layer of complexity: uncertainty is no longer a property of a single output but
something that accumulates across interdependent steps~\citep{zhang2025htc, duan2025uprop, zhao2024saup}.
HTC~\citep{zhang2025htc} confronts this directly, extracting hand-crafted trajectory features
from an agent's full log-probability sequence and fitting a logistic-regression calibrator. We
adopt HTC's trajectory-feature view and depart from it in two respects that prove decisive for
deployment: we separate reasoning from action spans, and we obtain the signal from a separate
draft model rather than the agent's model own logits.

\paragraph{Hallucinations and uncertainty in tool-using agents.}
A line of recent work probes confidence in tool-using agents.
\citet{internal_tool_selection, mice_for_cats} train probes on a target model's internal states to anticipate erroneous
tool calls. Agents that reason before acting hallucinate tool calls more often~\citep{yin2025reasoning}. While those methods are effective, but operated on white-box access. \citet{confidence_dichotomy, agentic_overconfidence} documents a
systematic gap between expressed and actual confidence in tool-use agents.  

\paragraph{Speculative decoding and draft models.}
Speculative decoding~\citep{leviathan2023fast, chen2023accelerating} pairs a small draft
model with a large target, accepting drafted tokens with alignment-specific probability. Its purpose is mainly to accelerate inference~\citep{xia2024unlocking, leviathan2023fast}. We
retain the machinery and invert the objective: the small model does not generate, it
\emph{evaluates}, scoring the target's committed tokens after the fact.

\section{Method}
\label{sec:method}

\subsection{Problem Setup}
\label{sec:setup}

Let $p$ be the large, possibly black-box \emph{agent model} (further simply the \textbf{agent}),
and $q$ a small open-weight \emph{draft model}. Given a task with initial
context $x$, the agent produces a trajectory of alternating reasoning and
action spans:
\begin{equation*}
  \mathcal{T} = \bigl(r_1,\, a_1,\, r_2,\, a_2,\, \ldots,\, r_N,\, a_N\bigr),
\end{equation*}
where $r_i \in \Sigma^*$ is the $i$-th reasoning span and $a_i \in \Sigma^*$ is
the $i$-th action span. Let $y = (y_1, \ldots, y_T)$ be the in-order
concatenation of all tokens; each carries a component label
$\tau_t \in \{\texttt{reason},\, \texttt{action}\}$, read off the harness's message structure: in OpenHands~\citep{wang2025openhands} the assistant's free-form
text forms the reasoning span and the subsequent structured tool call (the
invoked function and its serialized arguments) forms the action span.

Each agent is paired with a \textbf{verifiable objective}: a binary function
$\Omega(a, x) \in \{0, 1\}$ that an external oracle evaluates on action $a$
given context $x$. SU operates at \emph{step} granularity: for each step $i$ it
predicts the oracle verdict $\Omega(a_i, x)$ on that step's action
\emph{before} $a_i$ executes. Rather than conditioning on the full
trajectory, which grows long and increasingly noisy over many steps, in this study we
condition on a fixed \emph{lookback window} of the $k=3$ most recent steps. Let
\begin{equation*}
  \mathcal{W}_i = \bigl(r_{i-k+1},\, a_{i-k+1},\, \ldots,\, r_{i-1},\, a_{i-1},\, r_i,\, a_i\bigr)
  \label{eq:window}
\end{equation*}
be the window ending at the step under evaluation (truncated at the start of
the trajectory when $i < k$). We seek a calibration function $\mathcal{F}$ that
maps the window to a calibrated probability that the step's action will pass its
check, \emph{before} $\Omega$ is evaluated:
\begin{gather*}
  \hat{y} = \mathcal{F}(\mathcal{W}_i) \in [0, 1], \\
  \mathbb{E}\bigl[\Omega(a_i,x) \mid \mathcal{F}(\mathcal{W}_i) = c\bigr] \approx c.
\end{gather*}
In our
deployment $\Omega$ is code-execution success, but the framework is agnostic to
the objective.

\subsection{Speculative Cross-Likelihoods}
\label{sec:cross_likelihood}

For each token $y_t$ we run the draft model $q$ in \emph{teacher-forcing mode}: conditioned on the
prefix $(x, y_{<t})$, we read off $q$'s next-token distribution and evaluate it at the token $p$
actually produced. A single forward pass over the trajectory yields three token-level signals at no
cost to the agent pipeline:
\begin{align*}
  s_t &= -\log q(y_t \mid x,\, y_{<t}), \\
  g_t &= \log q(\hat{y}_t \mid x,\, y_{<t}) - \log q(y_t \mid x,\, y_{<t}), \\
  H_t &= -\!\sum_{v \in V} q(v \mid x,\, y_{<t})\,\log q(v \mid x,\, y_{<t}),
\end{align*}
where $\hat{y}_t = \argmax_v\, q(v \mid x, y_{<t})$. We call $s_t$ the \emph{speculative
surprisal}, $g_t$ the \emph{speculative gap}, and $H_t$ the \emph{speculative entropy}. All three
depend only on the observed token stream; $p$'s logits are never needed at inference.

\paragraph{Why a black-box proxy works.}
In speculative decoding, a drafted token is accepted with probability
$\alpha_t = \min(1, p(y_t|{\cdot})/q(y_t|{\cdot}))$. When $p$ is a black box, $p(y_t|{\cdot})$ is
hidden, therefor if $q$ tracks $p$ reasonably well, a high speculative surprisal $s_t$ implies $p$ too
would face a low acceptance rate there, making $s_t$ a usable proxy for $p$'s own uncertainty. This is why aligning $q$ to $p$ matters.

\subsection{Component-Aware Feature Extraction}
\label{sec:features}

An agentic trajectory alternates between a high-entropy \textbf{reasoning} span
and a sharply lower-entropy \textbf{action} span, so the speculative signals
carry phase-dependent meaning (we analyze this phase structure in
\autoref{app:dynamics}). We therefore
compute features separately for the reasoning and action tokens within the
lookback window $\mathcal{W}_i$.
For each of the
two phases (reasoning, action) and each of the three signals ($s$, $g$, $H$),
we summarize the corresponding tokens with eight statistics---mean, variance,
max, min, skewness, trend, and the mean over the first and last 10\% of
tokens, giving 48 features in total. Adding the two span lengths
yields the full feature vector $\phi(\mathcal{W}_i)\in\mathbb{R}^{50}$ (full
taxonomy in \autoref{app:features}). Computing the phases separately keeps the
noisy reasoning signal from masking the small, failure-relevant deviations in
the action span, and the short window keeps those deviations from being noised by earlier steps.

\subsection{Calibrator and Downstream Policy}
\label{sec:calibrator}

Following HTC~\citep{zhang2025htc} framework, we fit an $\ell_1$-regularized logistic
regression on labeled windows. With given features
$\phi(\mathcal{W}_i) \in \mathbb{R}^{50}$ and binary labels $y \in \{0,1\}$,
the calibrator $\mathcal{F}(\phi) = \sigma(w^\top \phi + b)$ is fit by minimizing
\begin{equation*}
  \min_{w, b}\ \sum_{j} \mathcal{L}\bigl(y_j,\, \sigma(w^\top \phi + b)\bigr)
  + \lambda \lVert w \rVert_1,
  \label{eq:logreg}
\end{equation*}
where $\mathcal{L}$ is the binary cross-entropy, $\sigma$ the logistic function,
and $\lambda$ the $\ell_1$ penalty that selects a sparse, auditable subset of
the 50 features. We deliberately keep the calibrator simple, since a linear model is
cheap to train and it is robust in the small-data regime typical of SWE benchmarks.

The output of SU is a single scalar failure-likelihood score: the estimated probability $\hat{y} =
\mathcal{F}^*(\phi(\mathcal{W}_i))$ that the next action will follow the verifiable objective.
What a system \emph{does} with that probability is a separate, application-level
decision, and the framework is deliberately agnostic to it. A calibrated
failure signal can drive many policies, for instance, routing the step to a larger
or more capable model when confidence is low, escalating to a human reviewer or
allocating extra test-time compute (e.g., resampling) to risky steps.

\paragraph{The veto gate.}
To make the deployment impact measurable, we study the policy we deploy and
evaluate in this work: a \emph{veto gate} inserted between code generation and
execution (\autoref{alg:veto}). When the predicted success probability falls
below a threshold $\tau$, the gate blocks the action and triggers a cheaper
replan instead of an execute-fail-retry cycle. We choose this policy because its
effect is transparent. It maps the probability to a single,
auditable accept-replan decision, so the error and cost reduction numbers in
\autoref{sec:impact} are directly attributable to the quality of the underlying
signal. 
The threshold $\tau$ trades true vetoes (blocking code that would fail) against false vetoes
(blocking code that would succeed). $\tau$ is selected on the training split and frozen for all reported evaluations. We emphasize that the gate is \emph{one} policy among many the signal could drive, not a requirement of the framework: any of the
alternatives described above could consume $\hat{y}$ in its place.

\begin{algorithm}[t]
\caption{SU Veto-Gate Policy}
\label{alg:veto}
\begin{algorithmic}[1]
  \Require Window $\mathcal{W}_i$, threshold $\tau$, max retries $K$
  \State Compute $\phi(\mathcal{W}_i)$ via draft teacher forcing \Comment{single pass, $O(|\mathcal{W}_i|)$}
  \State $\hat{y} \leftarrow \mathcal{F}^*(\phi(\mathcal{W}_i))$ \Comment{calibrated $P(\text{success})$}
  \If{$\hat{y} \geq \tau$}
    \State \textbf{execute:} send action $a_i$ to the environment
  \ElsIf{retries $< K$}
    \State \textbf{replan:} append a failure hint and re-invoke the agent
  \Else
    \State \textbf{abstain:} return with a low-confidence flag
  \EndIf
\end{algorithmic}
\end{algorithm}

\section{Experimental Results}
\label{sec:experiments}

We evaluate SU in agentic software engineering tasks, where $\Omega$ is whether generated code runs without exceptions: action spans are long, syntactically and contextually constrained, and the objective is binary verifiable feedback from the execution environment.

\paragraph{Setup.}
We train both draft model and calibrator on the SWE-rebench OpenHands
trajectories~\citep{trofimova2025openhandstrajs}, real-world GitHub issues
from SWE-rebench~\citep{badertdinov2026swe} solved by Qwen3-Coder-480B under the
OpenHands scaffold~\citep{wang2025openhands}. Further, we evaluate out-of-distribution on a
held-out split, on SWE-Bench
Verified~\citep{jimenez2024swebench}, and on DA-Code~\citep{dacode}.
Success is scored using the execution data derived from the agentic trajectories. We use
Qwen3-Coder-480B~\citep{qwen3coder} and the fully closed-source Claude~3.5~Sonnet~\citep{claude35} as agent models ($p$), and Qwen3-4B~\citep{yang2025qwen3} as a draft model ($q$). In the study we validate base model, SFT-tuned model, and teacher-forced (TF) distilled variants.
As a baselines we use Verbalized Confidence~\citep{tian2023just} made by the agent; Last-TP and
Global-TP, mean log-probability of the final block and whole trajectory respectively under $p$ counting it as a
white-box; and HTC~\citep{zhang2025htc} on agent's own log-probabilities. We ablate SU-Uniform, our signals without phase separation. We report
AUROC, Precision at Recall 80 (P@R80) calibration metrics. For the gate mechanism evaluation we additionally report true-/false-veto rates and net
error/cost change. Implementation details, training recipes, and compute budgets are described in
\autoref{app:repro}.

\subsection{Failure-Prediction Quality}
\label{sec:quality}

\autoref{tab:main_results} shows
failure-prediction quality across the three splits. First, \textbf{importance of alignment}: an untrained draft aligns poorly with the
agent and is barely above verbalized confidence (\Hnumber{.61} AUROC on SWE-Bench Verified), while
distillation lifts SU to \Hnumber{.77}, and both SFT and TF land within \Hnumber{1}. Second, phase separation is worth \Hnumber{+5--6} AUROC:
dropping it (SU-Uniform, \Hnumber{.71}) confirms the action-span signal becomes worse once phases are
pooled. 
Third, SU recovers most of the white-box ceiling: it lifts AUROC from \Hnumber{.57} (verbalized confidence) to \Hnumber{.77}, approaching HTC (\Hnumber{.82}), using only output tokens and showing improvements over
the white-box log-probability baselines (Last-/Global-TP). It does not exceed HTC, and not expected to:
a black-box proxy approaching its white-box ceiling is the credible outcome. 

\begin{table}[t]
\centering
\caption{Code-execution failure prediction (agent: Qwen3-Coder-480B; SU uses Qwen3-4B draft).
$\dagger$~In-distribution held-out split; SWE-Bench Verified shares the agent and scaffold and differs only in task distribution; DA-Code additionally changes domain and tool interface.}
\label{tab:main_results}
\footnotesize
\setlength{\tabcolsep}{4pt}
\renewcommand{\arraystretch}{1.05}
\resizebox{\columnwidth}{!}{%
\begin{tabular}{@{}lcccccc@{}}
\toprule
& \multicolumn{2}{c}{\textbf{SWE-rebench}$^\dagger$}
& \multicolumn{2}{c}{\textbf{SWE-Bench Ver.}}
& \multicolumn{2}{c}{\textbf{DA-Code}} \\
\cmidrule(lr){2-3}\cmidrule(lr){4-5}\cmidrule(lr){6-7}
\textbf{Method}
  & \textbf{AUC} & \textbf{P@R80}
  & \textbf{AUC} & \textbf{P@R80}
  & \textbf{AUC} & \textbf{P@R80} \\
\midrule
Verbalized Conf.
  & \Hnumber{.58} & \Hnumber{.50} & \Hnumber{.57} & \Hnumber{.49} & \Hnumber{.59} & \Hnumber{.51} \\
\midrule
Last-TP
  & \Hnumber{.70} & \Hnumber{.61} & \Hnumber{.68} & \Hnumber{.59} & \Hnumber{.69} & \Hnumber{.60} \\
Global-TP
  & \Hnumber{.67} & \Hnumber{.57} & \Hnumber{.65} & \Hnumber{.55} & \Hnumber{.66} & \Hnumber{.56} \\
HTC (Agent)
  & \Hnumber{.85} & \Hnumber{.77} & \Hnumber{.82} & \Hnumber{.74} & \Hnumber{.83} & \Hnumber{.75} \\
\midrule
SU, no train
  & \Hnumber{.63} & \Hnumber{.54} & \Hnumber{.61} & \Hnumber{.52} & \Hnumber{.63} & \Hnumber{.54} \\
SU-Uniform (SFT)
  & \Hnumber{.74} & \Hnumber{.65} & \Hnumber{.71} & \Hnumber{.62} & \Hnumber{.72} & \Hnumber{.63} \\
\rowcolor{gray!12}
\textbf{SU, SFT}
  & \Hnumber{.79} & \Hnumber{.70}
  & \Hnumber{.76} & \Hnumber{.67}
  & \Hnumber{.77} & \Hnumber{.68} \\
\rowcolor{gray!12}
\textbf{SU, TF}
  & \Hnumber{.80} & \Hnumber{.71}
  & \Hnumber{.77} & \Hnumber{.68}
  & \Hnumber{.78} & \Hnumber{.69} \\
\bottomrule
\end{tabular}%
}
\end{table}

\begin{table}[t]
\centering
\caption{Deployment impact of the SU veto gate (Qwen3-Coder-480B agent, Qwen3-4B draft).
SWE-Bench Verified and DA-Code are evaluated without additional fine-tuning.}
\label{tab:impact}
\small
\setlength{\tabcolsep}{4pt}
\renewcommand{\arraystretch}{1.1}
\begin{tabular}{@{}lcc@{}}
\toprule
& \textbf{SWE-Bench Verified} & \textbf{DA-Code} \\
\midrule
No-veto error rate     & \Hnumber{\impNoVetoErrSB}\%          & \Hnumber{\impNoVetoErrDA}\% \\
SU-gated error rate    & \Hnumber{\impGatedErrSB}\%           & \Hnumber{\impGatedErrDA}\% \\
$\Delta$ error         & \textbf{\Hnumber{$-$\impDeltaErrSB}~pp} & \textbf{\Hnumber{$-$\impDeltaErrDA}~pp} \\
Token savings / task   & \textbf{\Hnumber{$-$\impTokSaveSB}\%}  & \textbf{\Hnumber{$-$\impTokSaveDA}\%} \\
\midrule
No-veto task success   & \Hnumber{\impNoVetoSuccSB}\%          & \Hnumber{\impNoVetoSuccDA}\% \\
SU-gated task success  & \Hnumber{\impGatedSuccSB}\%          & \Hnumber{\impGatedSuccDA}\% \\
$\Delta$ success        & \Hnumber{$-$\impDeltaSuccSB}~\%        & \Hnumber{$-$\impDeltaSuccDA}~\% \\
\midrule
True-veto recall       & \Hnumber{\impRecallSB}          & \Hnumber{\impRecallDA} \\
False-veto rate        & \Hnumber{\impFvrSB}          & \Hnumber{\impFvrDA} \\
\bottomrule
\end{tabular}
\end{table}

\subsection{Calibration}
\label{sec:calibration}

Discrimination and calibration are distinct properties, and SU delivers them
unequally. \autoref{tab:calibration} reports proper-scoring and calibration
metrics on the held-out SWE-rebench split against a constant predictor that
always emits the base rate. Alignment is what
produces utility: distillation improves the Brier score from
\Hnumber{.1265} to \Hnumber{.1012} and the Brier skill score from
\Hnumber{.019} to \Hnumber{.215} over the constant reference, consistent with
the AUROC gains of \autoref{sec:quality}.

We therefore describe SU's raw output as a failure-likelihood score
rather than a calibrated probability, and note that the veto gate of
\autoref{sec:impact} requires only a monotone score plus a threshold, so its
results depend on ranking quality alone and are unaffected by the residual
miscalibration. Standard post-hoc recalibration, such as temperature scaling or isotonic
regression is order-preserving and would leave AUROC and the gate unchanged
while reducing ECE. We did not apply it here, and we flag closing the
reliability gap as required work before SU's output is consumed by any policy that reads the probability value itself rather than thresholding it.

\begin{table}[t]
\centering
\caption{Calibration and proper-scoring metrics on the held-out SWE-rebench
split (agent: Qwen3-Coder-480B, Qwen3-4B draft). The constant reference always
predicts the base rate, it is trivially reliable and carries no information.}
\label{tab:calibration}
\footnotesize
\setlength{\tabcolsep}{4pt}
\renewcommand{\arraystretch}{1.05}
\resizebox{\columnwidth}{!}{%
\begin{tabular}{@{}lccccc@{}}
\toprule
\textbf{Method}
  & \textbf{ECE}$\downarrow$ & \textbf{MCE}$\downarrow$
  & \textbf{Brier}$\downarrow$ & \textbf{NLL}$\downarrow$
  & \textbf{BSS}$\uparrow$ \\
\midrule
Constant (base rate)
  & \Hnumber{.008} & \Hnumber{.008} & \Hnumber{.129} & \Hnumber{.426} & \Hnumber{.000} \\
SU, no train
  & \Hnumber{.015} & \Hnumber{.054} & \Hnumber{.127} & \Hnumber{.422} & \Hnumber{.019} \\
\rowcolor{gray!12}
\textbf{SU, SFT}
  & \Hnumber{.024} & \Hnumber{.060} & \Hnumber{.101} & \Hnumber{.364} & \Hnumber{.215} \\
\bottomrule
\end{tabular}%
}
\end{table}

\subsection{Deployment Impact: Fewer Errors, Lower Cost}
\label{sec:impact}

We lead with the result that matters most for deployment: does the gate (\autoref{alg:veto})
actually change how the agent behaves on the two axes a practitioner is accountable for---error
rate and compute? \autoref{tab:impact} shows that it does.

\paragraph{Error rate.}
At the operating threshold, SU cuts the per-call execution error rate from \impNoVetoErrSB\% to \impGatedErrSB\% on SWE-Bench Verified and from \impNoVetoErrDA\% to \impGatedErrDA\% on DA-Code. Crucially, the DA-Code
result was obtained without additional fine-tuning: the calibrator was trained on the SWE-rebench trajectories and applied without retraining.

\paragraph{Compute.}
Each failing trajectory, left alone, triggers a retry. By catching failures before execution and
swapping the full retry loop for a cheaper replan, the gate cuts average tokens per task by
\impTokSaveSB--\impTokSaveDA\% on both benchmarks. At production volume this is a direct reduction in API spend and
latency.

\paragraph{Task success rate.}
A natural question is whether cutting errors before execution also lifts
end-to-end task success. On
SWE-Bench Verified the resolved rate moves slightly down, from
\impNoVetoSuccSB\% to \impGatedSuccSB\%, and on DA-Code it is essentially unchanged
(\impNoVetoSuccDA\% to \impGatedSuccDA\%). The reason is structural: the veto gate
suppresses confidently-wrong actions and swaps the execute-fail-retry loop
for a cheaper replan, but a replan is not guaranteed to recover the task,
and a small fraction of false vetoes block actions that would have
succeeded. The deployment value of SU therefore lies in the cost and
silent-failure axes, fewer wasted execute-fail-retry cycles and
\impTokSaveSB--\impTokSaveDA\% lower token spend, rather than in raising the benchmark
scores. Raising task success would require a downstream policy stronger than the simple replan we study here, which we leave as a promising direction for
further research.

\subsection{Cross-Agent Transfer}
\label{sec:cross_agent}

A deployment-critical question is whether a trained draft generalizes to agents
it was not specifically aligned with, since teams frequently serve several
agents behind a single interface. We examine this through several training
configurations, summarized in \autoref{tab:cross_agent}.

When a strong open-weight agent is available, the draft can be aligned to it via
SFT or TF distillation. However, the two
objectives yield closely matched scores (\Hnumber{.76} vs. \Hnumber{.77} on Qwen),
showing that while the alignment is crucial, the particular
distillation method is not as important. More importantly, a draft aligned to one agent still
transfers to a different agent: a Qwen-distilled draft reaches
\Hnumber{.69} AUROC on Claude~3.5~Sonnet, well above the untrained baseline
(\Hnumber{.60}). We attribute this to two factors. First, fine-tuning on agent
trajectories improves alignment to the software engineering domain rather than
to the specific agent alone. Second, the adapted draft is less surprised by
in-domain trajectories than the base model, so its speculative signal remains
informative even when the target agent differs.

Most notably, training on the open-weight corpus alone is close to and
competitive with mixed-corpus training: a Qwen-distilled draft reaches
\Hnumber{.69} on Claude, within a few points of the mixed-corpus result
(\Hnumber{.75}). This is promising, as it suggests that an open, logit-accessible
agent can serve as an effective proxy for aligning drafts toward closed-source
targets, motivating the post-training methods that exploit logit access or hidden representation on
the open model while still transferring to black-box agents.

\begin{table}[t]
\centering
\caption{Cross-agent transfer on SWE-Bench Verified (AUROC / P@R80; Qwen3-4B draft). 
Claude~3.5~Sonnet is fully black-box (no logits).}
\label{tab:cross_agent}
\small
\setlength{\tabcolsep}{5pt}
\renewcommand{\arraystretch}{1.1}
\begin{tabular}{@{}ccc@{}}
\toprule
\textbf{Train corpus}
  & \textbf{Qwen3-480B} & \textbf{Claude~3.5~Sonnet} \\
\midrule
---        & \Hnumber{.61} / \Hnumber{.52} & \Hnumber{.60} / \Hnumber{.52} \\
Qwen (TF)   & \Hnumber{.77} / \Hnumber{.68} & \Hnumber{.69} / \Hnumber{.61} \\
Qwen (SFT)   & \Hnumber{.76} / \Hnumber{.67} & \Hnumber{.69} / \Hnumber{.60} \\
Claude (SFT) & \Hnumber{.69} / \Hnumber{.61} & \Hnumber{.76} / \Hnumber{.67} \\
Mixed (SFT)  & \Hnumber{.76} / \Hnumber{.67} & \Hnumber{.75} / \Hnumber{.66} \\
\bottomrule
\end{tabular}
\end{table}

\section{Discussion and Practical Takeaways}
\label{sec:discussion}

\paragraph{The deployment recipe.}
For teams running frontier API agents: pair the agent with a distilled \Hnumber{4B}
open-weight draft, distill it on a few thousands labeled trajectories
from the target agent (or a mixed corpus if several agents are in play), fit a linear calibrator
with phase-aware features, and insert the veto gate before execution. The whole pipeline adds one
small forward pass per step and no changes to the agent, and in our study returns \impDeltaErrSB--\impDeltaErrDA~pp
lower error and \impTokSaveSB--\impTokSaveDA\% lower token cost.

\paragraph{Transfer beyond code.}
SU needs only two things from a domain: trajectories with identifiable reasoning and action spans,
and an external binary oracle over actions. Both hold beyond software engineering, such as SQL agents
($\Omega$ checks query results), web agents ($\Omega$ checks a target state), and
tool-orchestration agents ($\Omega$ checks the right call with the right arguments). In each, the
deliberation-then-commitment structure persists, and with it the entropy divergence SU relies on.
We have not measured these settings and make no quantitative claim about them; confirming the
transfer is the natural next step.

\paragraph{From observation to intervention.}
The veto gate is not a passive monitor but an excplicit intervention: it reshapes the trajectory by
injecting a replanning prompt exactly where failure is predicted. A natural extension is a
\emph{streaming} veto that watches the action span's entropy token by token and interrupts the
instant it diverges and catching failure mid-generation, before the block is finished. We leave this
to future work.

\section{Conclusion}
\label{sec:conclusion}

Production agents fail confidently and expensively, paying for each erroneous action before its consequences reveal the error. \textbf{Speculative Uncertainty} anticipates the failure instead: a small open-weight
draft model scores a black-box agent's trajectory in a single forward pass, and a linear calibrator
turns the result into a failure-likelihood score that gates execution. Deployed this way it cut the per-call execution error rate by \Hnumber{6--8} percentage points and token cost by \Hnumber{14--19}\%, with gains that transfer
zero-shot and extend to a fully closed-source agent.
Two design choices enable this:
scoring the agent through a separate model rather than its internals, and respecting the
reasoning--action phase structure of generation. We expect both to transfer to other
verifiable-objective settings in which an agent reasons its way to a committed action. Where that
structure recurs, an agent's failure can be anticipated rather than merely discovered.

\section*{Limitations}

\paragraph{Quality of resulted models.}
SU consists of several interacting components, each of which influences final
performance: the choice of draft model and its size, the
alignment procedure, the feature
construction, and the calibrator training. The combinatorial space these define is large,
and we do not claim to identify its optimum. Our contribution is instead a
lightweight and scalable framework together with empirical guidance on
reasonable operating points. Tuning any individual component further is likely to yield additional gains we do not pursue here.

\paragraph{A single objective and domain.}
Our evidence is drawn entirely from code-execution success as the verifiable
objective. While the framework itself is agnostic to the objective---requiring
only trajectories with identifiable reasoning and action spans and an external
binary oracle---we provide no quantitative evidence for other settings.
Extending SU to additional domains and objectives is an important direction for future work.

\paragraph{Single-run estimates.} All reported numbers are single-run point estimates on fixed evaluation sets; we do not report seed variance or confidence intervals. Differences of a few AUROC points, and the 5 percentage points SWE-Bench Verified task-success change, should be read with that in mind.

\paragraph{The draft cost is objective-dependent.}
Throughout, we treat the draft's per-step forward pass as effectively free
relative to the agent and the avoided retries, which holds for the code-execution
setting we study. This assumption may not transfer to objectives that demand a
larger or more capable draft to produce an informative speculative signal, in
which case the per-step overhead must be re-evaluated against the downstream
savings. The favorable cost balance we report is therefore specific to our
workload and should not be assumed to hold universally.

\section*{Acknowledgments}

We acknowledge the use of LLMs for text polishing and language improvements throughout this manuscript. All technical content, ideas, and substantial writing remain the original work of the authors.

\bibliography{references}

\appendix

\newpage
\section{The Phase Structure in Tool Augmented Agents}
\label{app:dynamics}

An \emph{agentic trajectory}
is the sequence of tokens an agent emits while solving a task. The exact format depends on the
harness, but the dominant pattern across modern agent frameworks based on the reasoning LLMs is roughly the same: within each step
the model first generates a \textbf{reasoning} span (planning, analysis) and then an
\textbf{action} span (the tool call or code to be executed). Although both come from a single
autoregressive pass, their token-level dynamics differ significantly.

\begin{figure*}[t]
  \centering
  \includegraphics[width=0.95\textwidth]{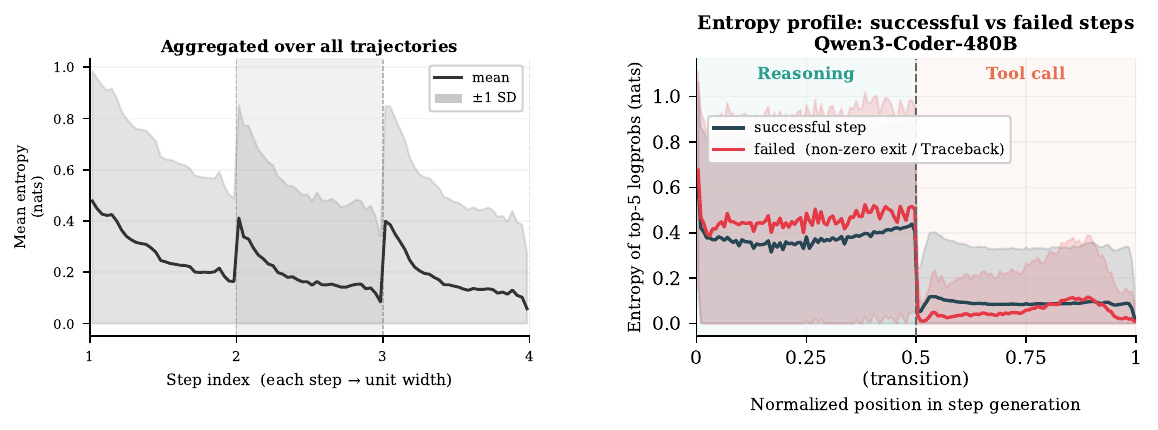}
  \caption{\textbf{Speculative entropy of agentic generations.}
    \emph{Left:} mean entropy aggregated over all steps---the per-step
    reasoning-to-action entropy drop is washed out and not visible.
    \emph{Right:} entropy along a normalized step (reasoning $x<0.5$, action
    $x>0.5$) for successful (navy) vs. failing (red) actions, revealing the
    sharp collapse at the transition. The failure signal appears in \emph{both}
    phases with opposite sign, failures are more uncertain during reasoning,
    less so in the action span, which is why phase-aware features help. Bands
    show $\pm1\sigma$ over \Hnumber{500} training trajectories.}
  \label{fig:entropy_profiles}
\end{figure*}

\autoref{fig:entropy_profiles} traces per-token entropy along agentic generations, with position
normalized so that $x=0$ marks the start of reasoning, $x=0.5$ marks the transition to the action
span, and $x=1$ marks its end, separately for successful and failing actions over \Hnumber{500}
SWE-rebench OpenHands trajectories (\Hnumber{250} success, \Hnumber{250} failure). Two empirical findings stand
out, and each shapes the method. First, reasoning is highly exploratory:
entropy is high as the model weighs alternatives, then \emph{collapses} sharply
at the transition. Second, and most important, the failure signal is
implicitly present in \emph{both} phases.

This drop of entropy has a direct, practical consequence. Since the failure signal
points one way during reasoning and the opposite way inside the action span,
any calibrator that combines the two phases into the single sequence of tokens does not merely represent the signal. Moreover, it actively \emph{cancels} it:
the elevated-reasoning and suppressed-action contributions offset each other.
This motivates the central design choice of our method: extract features from
the reasoning and action spans \emph{separately}.

The phase structure in \autoref{fig:entropy_profiles} is shown for a single
step; in practice we aggregate these per-phase statistics over a short window of
recent steps (\autoref{sec:setup}) to give the calibrator local trajectory
context without the noise of the full history.

\section{Reproducibility: Data, Training, and Compute}
\label{app:repro}

This appendix documents every step needed to reproduce the results in
\autoref{sec:experiments}. We organize it as the pipeline runs: trajectory collection and
labeling, draft-model training, calibrator fitting, and the gate simulation.

\subsection{Trajectory Collection and Labeling}
All trajectories are collected under a single agentic scaffold, OpenHands~\citep{wang2025openhands}, so that
train and test distributions differ only in task content, not in formatting or tool interface;
this isolates genuine distribution shift from harness artifacts. For each task the agent emits a
sequence of reasoning and action spans: we map the assistant message's free-form
text to the reasoning span and its structured tool call (the invoked function and
its serialized arguments) to the action span, which we parse directly into the component
labels $\tau_t$ of \autoref{sec:setup}. Each action step is labeled by its execution outcome, the
oracle verdict $\Omega(a_i,x)$ the gate predicts (\autoref{sec:setup}): \Hnumber{1} if the tool call
executes without error (exit code \Hnumber{0}), \Hnumber{0} otherwise. We took SWE-rebench OpenHands
trajectories, which were
generated by Qwen3-Coder-480B. We use \Hnumber{10{,}000} of these trajectories for training and a
held-out \Hnumber{1{,}000} for testing, plus \Hnumber{500} SWE-Bench Verified and \Hnumber{500} DA-Code
trajectories used \emph{only} for out-of-distribution evaluation. These trajectories are long-horizon:
\Hnumber{$\sim$65} agent turns and \Hnumber{$\sim$54} thousands tokens of context on average. Across the training split this totals \Hnumber{$\sim$0.54}B context tokens for training.
\autoref{tab:data_stats} summarizes the splits and their base tool-call success rates, this rate is
the positive-class prevalence the calibrator must handle.

\begin{table}[h]
\centering
\caption{Dataset splits and base tool-call success rate (Qwen3-Coder-480B): the fraction of action
steps whose tool call executes without error (exit code \Hnumber{0}). Its complement is the base
per-call execution-error rate.}
\label{tab:data_stats}
\small
\setlength{\tabcolsep}{5pt}
\begin{tabular}{@{}lccc@{}}
\toprule
\textbf{Dataset} & \textbf{\#Traj.} & \textbf{\#Calls} & \textbf{Succ. \%} \\
\midrule
SWE-rebench (train) & \Hnumber{10{,}000} & \Hnumber{327{,}098} & \Hnumber{84.7} \\
SWE-rebench (held-out) & \Hnumber{1{,}000}  & \Hnumber{32{,}585} & \Hnumber{84.7} \\
\bottomrule
\end{tabular}
\end{table}

\subsection{Draft-Model Training}
We compare three regimes for each draft size. \textbf{No training} uses the released Qwen3-4B
checkpoint as-is. \textbf{SFT} fine-tunes the draft to reproduce the agent's trajectories with a
standard next-token cross-entropy loss on the agent's emitted tokens. \textbf{Teacher-forced (TF)
distillation} additionally aligns the draft's full next-token distribution to the agent's. We keep
training-time and inference-time access separate: at inference SU is strictly black-box, reading
only the agent's realized output tokens, whereas draft training is a one-time offline step. Because
the agent Qwen3-Coder-480B is open-weight, we recover its logits in a single forward pass over the
collected trajectories and distill the smaller draft against this full distribution. This
soft-distribution supervision yields only marginal gains over realized-token SFT
(\autoref{tab:main_results}), consistent with our finding (\autoref{sec:cross_agent}) that the training
corpus matters more than the distillation objective. The fully closed Claude~3.5~Sonnet exposes no
logits and therefore cannot be TF-distilled; the draft used there is aligned by SFT on realized
tokens alone.
\autoref{tab:hparams} shows hyperparameters, which were used for draft model training.

\begin{table}[h]
\centering
\caption{Draft-model fine-tuning hyperparameters, held fixed across sizes.}
\label{tab:hparams}
\small
\setlength{\tabcolsep}{5pt}
\begin{tabular}{@{}ll@{}}
\toprule
\textbf{Hyperparameter} & \textbf{Value} \\
\midrule
Base model           & \Hnumber{Qwen3-4B} \\
Optimizer            & \Hnumber{AdamW} ($\beta_1{=}0.9,\ \beta_2{=}0.95$) \\
Learning rate        & \Hnumber{1e-5} (cosine, \Hnumber{5}\% warmup) \\
Weight decay         & \Hnumber{0.1} \\
Gradient clipping    & \Hnumber{1.0} \\
Epochs               & \Hnumber{2} \\
Effective batch size & \Hnumber{32} sequences \\
Max sequence length  & \Hnumber{32768} tokens \\
Context extension    & \Hnumber{YaRN} \\
Loss masking         & \Hnumber{assistant tokens only} \\
Precision            & \Hnumber{bf16} \\
LoRA / full FT       & \Hnumber{full fine-tuning} \\
\bottomrule
\end{tabular}
\end{table}

\subsection{Calibrator Fitting}
The calibrator is an $\ell_1$-regularized logistic regression over the
$\phi(\mathcal{W}_i)\in\mathbb{R}^{50}$ features of \autoref{app:features}. Features are standardized
(zero mean, unit variance) using statistics computed on the training split only, to avoid leakage
into the OOD evaluations. The regularization strength $\lambda$ is selected by 5-fold
cross-validation on the SWE-rebench training split over the grid \Hnumber{$\{10^{-3},\dots,10^{1}\}$},
optimizing validation AUROC; the selected value is then frozen for all test
evaluations, including the zero-shot DA-Code transfer.

\subsection{Compute and Latency}
\autoref{tab:compute} reports the cost of each pipeline stage. The figure most relevant to
deployment is the per-step draft latency, since it is the only overhead SU adds on the critical
path: a single teacher-forced forward pass over the trajectory-so-far, with no generation. For the \Hnumber{4B} draft model this is \Hnumber{$\approx$100}\,ms per step on a single A100 GPU, against much larger latencies of agent model, so the gate's overhead is a small fraction of a step.

\begin{table}[h]
\centering
\caption{Compute budget by pipeline stage. Per-step latency is the only overhead on the agent's
critical path.}
\label{tab:compute}
\small
\setlength{\tabcolsep}{4pt}
\begin{tabular}{@{}lll@{}}
\toprule
\textbf{Stage} & \textbf{Hardware} & \textbf{Cost} \\
\midrule
Draft SFT/TF (4B)   & \Hnumber{4$\times$A100} & \Hnumber{$\sim$50}\,GPU-h \\
Calibrator fit      & \Hnumber{64$\times$CPU}  & \Hnumber{$<$1}\,min \\
Draft scoring (per step, 4B) & \Hnumber{1$\times$A100} & \Hnumber{$\approx$100}\,ms \\
\bottomrule
\end{tabular}
\end{table}

\section{Feature Taxonomy}
\label{app:features}

\autoref{tab:features} enumerates all \Hnumber{50} features in $\phi(\mathcal{T})$. The construction is
deliberately uniform: for each of the two component types ($\mathrm{R}$\,=\,reasoning,
$\mathrm{A}$\,=\,action) and each of the three speculative signals ($s$\,=\,surprisal, $g$\,=\,gap,
$H$\,=\,entropy), we compute the same eight aggregate statistics over the corresponding token subset,
giving $2\times3\times8=48$ statistical features; the final two are the reasoning- and action-span
lengths. The eight statistics are chosen to capture complementary aspects of a signal's
\emph{distribution within a span}: location (mean), spread (variance), extremes (max, min), asymmetry
(skewness), drift (trend $\Delta=\mu_\text{last}-\mu_\text{first}$), and boundary behavior (mean over the
first/last 10\% of tokens). The boundary and trend statistics are what let a \emph{linear} calibrator pick
up the late action-span entropy uptick of \autoref{fig:entropy_profiles}: that uptick raises
$\mu_\text{last}$ and the trend $\Delta$ specifically within the action component, while leaving the
reasoning component untouched, which is only possible because the two components are featurized
separately (\autoref{sec:features}).

\begin{table}[h]
\centering
\caption{Feature taxonomy for $\phi(\mathcal{T}) \in \mathbb{R}^{50}$.}
\label{tab:features}
\small
\setlength{\tabcolsep}{6pt}
\renewcommand{\arraystretch}{1.2}
\begin{tabular}{@{}llc@{}}
\toprule
\textbf{Group} & \textbf{Signal} & \textbf{\# stats} \\
\midrule
\multirow{3}{*}{Reasoning}
  & Surprisal $s$ & 8 \\
  & Gap $g$       & 8 \\
  & Entropy $H$   & 8 \\
\midrule
\multirow{3}{*}{Action}
  & Surprisal $s$ & 8 \\
  & Gap $g$       & 8 \\
  & Entropy $H$   & 8 \\
\midrule
Structural & $|\mathcal{I}^\mathrm{R}|,\ |\mathcal{I}^\mathrm{A}|$ & 2 \\
\bottomrule
\end{tabular}
\end{table}

\end{document}